\documentclass[letterpaper, 10 pt, conference]{ieeeconf} 
\usepackage{amsmath,amsfonts}
\usepackage{algorithmic}
\usepackage{algorithm}
\usepackage{array}
\usepackage[caption=false,font=normalsize,labelfont=sf,textfont=sf]{subfig}
\usepackage{textcomp}
\usepackage{stfloats}
\usepackage{url}
\usepackage{verbatim}
\usepackage{graphicx}
\usepackage{cite}
\usepackage{nomencl}
\usepackage{ntheorem}   
\usepackage{lipsum}     
\usepackage{graphicx}

\usepackage{xcolor}
\usepackage{hyperref}

\hypersetup{
	colorlinks=true,
	linkcolor=blue,
	filecolor=green, 
	urlcolor=black,
	citecolor=orange
}

\usepackage{bm}

\DeclareRobustCommand{\uvec}[1]{{%
		\ifcsname uvec#1\endcsname
		\csname uvec#1\endcsname
		\else
		\bm{\mathbf{#1}}%
		\fi
}}

\IEEEoverridecommandlockouts                              

\usepackage{graphicx} 

\title{\LARGE \bf
Novel Non-Prehensile Rolling Problem: Modelling and Balance Control of Pendulum-Driven Reconfigurable Disks Motion with Magnetic Coupling in Simulation
}

\author{Ollie Wiltshire and Seyed Amir Tafrishi*
\thanks{Ollie Wiltshire and Seyed Amir Tafrishi are with Geometric Mechanics and Mechatronics (gm$^2$R) Lab, the School of Engineering, Cardiff University, Cardiff, CF24 3AA, United Kingdom.
        {\tt\small \{wiltshireoj, tafrishisa\}@cardiff.ac.uk}}
\thanks{$^*$ Seyed Amir Tafrishi is the corresponding author.}}
\renewcommand{\baselinestretch}{.96}
\begin{document}

\maketitle
\thispagestyle{empty}
\pagestyle{empty}


\begin{abstract}
This paper presents a novel type of mobile rolling robot designed as a modular platform for non-prehensile manipulation, highlighting the associated control challenges in achieving balancing control of the robotic system. The developed rolling disk modules incorporate an innovative internally actuated magnetic-pendulum coupling mechanism, which introduces a compelling control problem due to the frictional and sliding interactions, as well as the magnetic effects between each module. In this paper, we derive the nonlinear dynamics of the robot using the Euler–Lagrange formulation. Then, through simulation, the motion behavior of the system is studied and analyzed, providing critical insights for future investigations into control methods for complex non-prehensile motion between robotic modules. Also, we study the balancing of this new platform and introduce a new motion pattern of lifting. This research aims to enhance the understanding and implementation of modular self-reconfigurable robots in various scenarios for future applications.
\end{abstract}

\section{Introduction}
Reconfigurable robots, composed of multiple modules that can interact and reconfigure, often lack versatility and independent movement. Coupling rolling robots as modular components addresses this issue by enabling displacement through internal actuation \cite{armour2006rolling,tafrishi2019design}. However, developing coupling mechanisms without external components remains a significant challenge \cite{seo2019modular,ruggiero2018nonprehensile}. Controlling these robots is complicated by the properties of non-prehensile manipulation, where modules can slip and lose their state during movement. Accurate control requires simulating and predicting these dynamics.

Modular self-reconfigurable robots (MSRRs) \cite{seo2019modular} are mobile robots with individual modules containing controllers, sensors, and actuation systems. These modules can reconfigure, enabling MSRRs to perform diverse tasks and navigate unpredictable environments \cite{981854, moubarak2012modular}. Modules connect via docking systems, such as mechanical surfaces, magnetic interactions, and grippers, including hybrid combinations. Reconfiguration can occur autonomously or manually. MSRRs are invaluable in search and rescue, environmental monitoring, and hazardous environments due to their modular and reconfigurable nature, which allows them to adjust their morphology for optimal stability and manoeuvrability \cite{moubarak2012modular, seo2019modular}. Utilizing a modular rolling robot at the centre of the reconfigurable system could be advantageous, as rolling motion offers an effective means of locomotion \cite{liang2020freebot}. Nevertheless, developing an appropriate coupling control method while creating locomotion between rolling modules remains an open challenge.
\begin{figure}[t!]
      \centering
      \includegraphics[width = 0.4\textwidth]{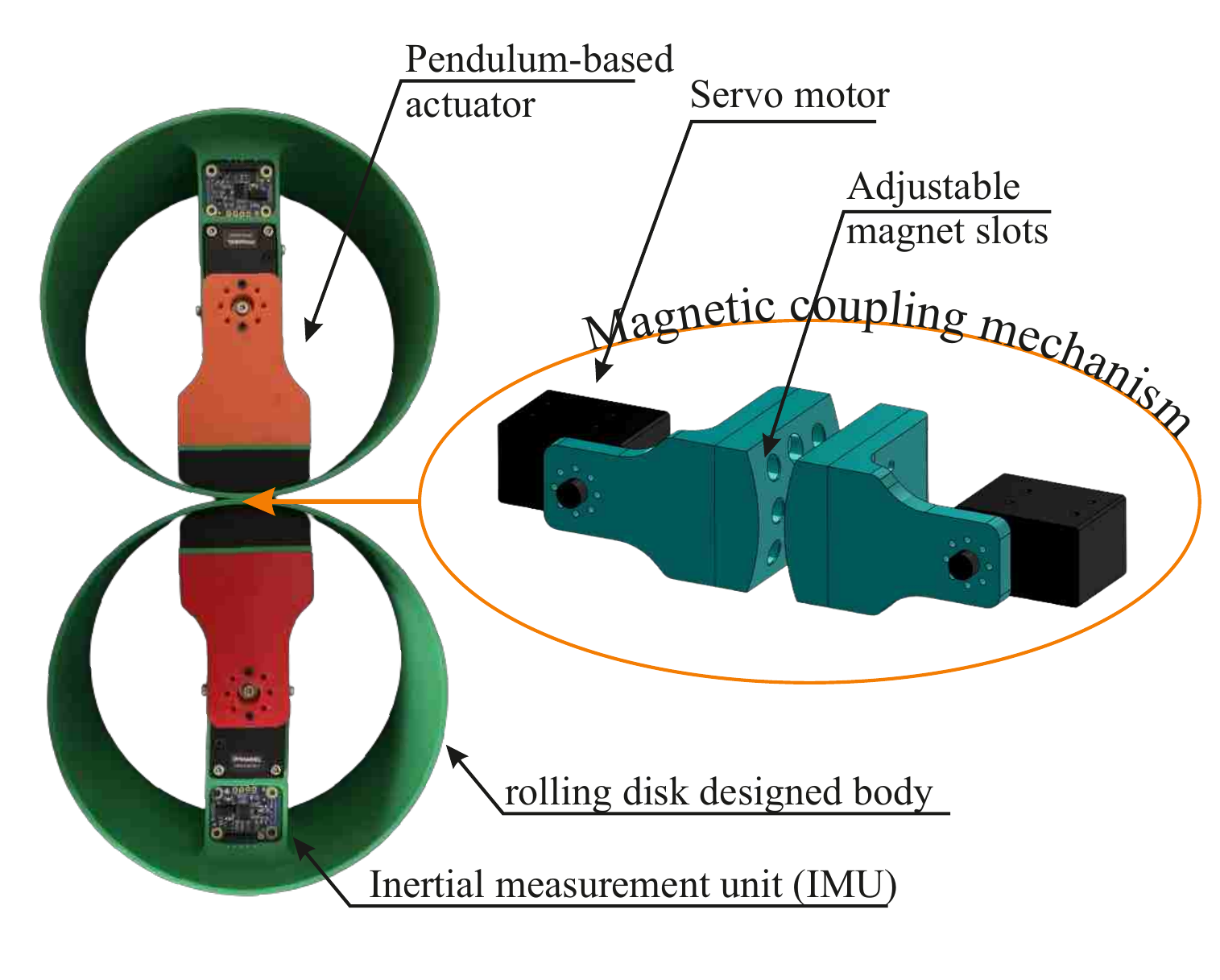}
      \caption{Reconfigurable Rolling Disk Robot design with magnetic coupling while balancing \cite{Ollie2024towards}}
      \label{fig:robot_design}
\end{figure}

During motion, rolling robots do not remain fixed to the surface and are subject to slippage, leading to a loss of state \cite{ruggiero2018nonprehensile}. This issue necessitates simulating and predicting the dynamics for accurate control, introducing new challenges in non-prehensile manipulation. Non-prehensile manipulation, characterized by the absence of direct grasping of objects \cite{mason1999progress,ruggiero2018nonprehensile}, presents significant control difficulties due to slippage and friction, resulting in unpredictable dynamics. Previous research has explored rolling and balancing disks \cite{6522171} and the control of a disk on a beam \cite{8613893}, highlighting the complexities involved in non-prehensile manipulation. A new modular design \cite{Ollie2024towards} as shown in Fig. \ref{fig:robot_design} addresses these challenges by introducing internally actuated modular disk modules, differing from existing approaches that rely on external actuation. This advancement opens great potential for applications in swarm reconfigurable robotics. However, due to the novelty of the locomotion in this new non-prehensile problem, the nonlinear models and capabilities of these types of systems are not yet well covered.

This paper derives the nonlinear dynamics using the Euler–Lagrange formulation in Section \ref{Sec:robordesign} for a novel internalized pendulum-based magnetic coupling mechanism for rolling reconfigurable disk robots, which is described briefly in Section \ref{Sec:modelinglagrangue}. Next, this work demonstrates the robot's distinctive motion capabilities using scenario-based PD controllers as an important contribution in Section \ref{sec:simulation}. Additionally, we introduce and discuss new nonprehensile manipulation problems, such as lifting, which are supported by simulation results that provide insights into the nonlinear dynamics of the system.


\begin{figure}[t!]
      \centering
       \includegraphics[width = 0.43\textwidth]{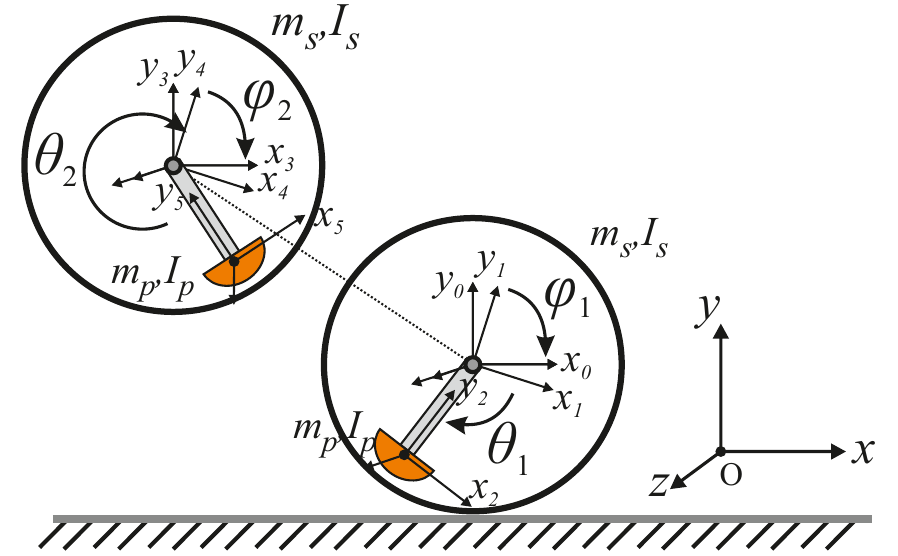}
      \caption{Schematic of the kinematic model.}\label{fig:Kinematics_Model_two_disks}
\end{figure}
 
\section{Reconfigurable Rolling Robot}\label{Sec:robordesign}
This section briefly showcases the design of the robot modules and the magnetic coupling mechanism as described in \cite{Ollie2024towards}. The modular robot's main body features a Dynamixel XM430 high-precision servo motor at the center, equipped with an IMU sensor to capture the orientation of the disk. Additionally, to enable the mobile disk to rotate independently or be coupled for dependent movement, the module is connected to a pendulum-mass with a magnetic docking system, as shown in Fig. \ref{fig:robot_design}. It is important to note that to demonstrate basic behaviors and analyze coupling, the model is simplified to a disk configuration with two independently operating modules, each utilizing a 1-DoF motor. Additional bodies can be incorporated by attaching passive magnets in various alignments or by adding extra motors to increase degrees of freedom (DoF). This platform focuses on studying motion behaviors such as coupling, spinning, and rolling due to friction, serving as a foundation for new platforms addressing non-prehensile manipulation problems \cite{ruggiero2018nonprehensile}, some of which will be explored in this work, including balancing and lifting.

To explore coupling connections using permanent magnets, we studied various magnetic configurations to determine the optimal distance and arrangement for achieving the strongest coupling. Additionally, during the decoupling process, the magnets facilitate successful detachment, as illustrated in Fig. \ref{fig:final arrangment}. We selected the final "H" array configuration with reversed polarity for the outer center magnets. This adjustment maximized magnetic flux at close distances while enabling easier decoupling, thanks to a consistent decrease in flux density with increasing distance.

\section{Robot Dynamic Modelling}\label{Sec:modelinglagrangue}
In this section, we introduce a novel two-rolling disk system with actuated pendulum kinematics. Using the Euler–Lagrange formulation, we then derive the nonlinear dynamics governing this system.

The current model of the robot consists of two rolling disk modules, as illustrated in Fig. \ref{fig:robot_design}. The frame $\uvec{O}$ represents the \textit{inertial frame}. The first rolling disk, which maintains contact with the ground, rotates by an angle $\varphi_1$ around the $z$-axis of the frame $\uvec{O}_1$ with respect to the inertial frame $\uvec{O}_0$. Additionally, the pendulum attached to the first disk rotates by an angle $\theta_1$ relative to the frame $\uvec{O}_1$. The second rolling disk, which primarily rolls on the surface of the first disk, rotates by an angle $\theta_2$ about $\varphi_1$ in the frame $\uvec{O}_4$ with respect to the frame $\uvec{O}_3$. Finally, the pendulum attached to the second disk rotates by an angle $\theta_2$ relative to the frame $\uvec{O}_4$.
\begin{table}[t!]
    \centering
    \caption{Parameters for the reconfigurable rolling disk robot}
    \label{tab:parameters}
    \begin{tabular}{c|c|c}
        \hline
        Parameter & Symbol & Value \\
        \hline
        Mass of pendulum & $m_p$ & 0.262 kg \\
        Mass of body & $m_s$ & 0.70 kg \\
        Inertia (pendulum) & $I_p$ & 0.1 kg·m² \\
        Inertia (body) & $I_s$ & 0.1 kg·m² \\
        Size (disk 1) & $r_1$ & 0.06 m \\
        Size (disk 2) & $r_2$ & 0.06 m \\
        Length (pendulum 1) & $R_1$ & 0.065 m \\
        Length (pendulum 2) & $R_2$ & 0.065 m \\
        Rotation (pendulum 1) & $\theta_1$ & - rad \\
        Rotation (pendulum 2) & $\theta_2$ & - rad\\
        Rotation (disk 1) & $\phi_1$ & - rad\\
        Rotation (disk 2) & $\phi_2$ & - rad\\
        \hline
    \end{tabular}
\end{table}

The relative positions of the bodies are defined with subscript $s_i =$ Body $p_i =$ Pendulum as shown in Fig. \ref{fig:Kinematics_Model_two_disks}. The parameters are broken down in Table \ref{tab:parameters}. The origin of the equation is defined from the centre of the first disk body at $\uvec{r_{s_1}}$.
\begin{align}
  &  \uvec{r_{s_1}} = \begin{bmatrix} R_1 \phi_1\\ 0\end{bmatrix}, \; \uvec{r}_{p_1} =     \begin{bmatrix}
        R_1 \phi_1+  r_1 \sin{(\phi_1 + \theta_1)} \\
        - r_1 \cos{(\phi_1 + \theta_1)}
    \end{bmatrix}  \nonumber\\
  &\uvec{r_{s_2}} = \uvec{r}_{s_1} + \uvec{r}_{s_2/s_1} =    
    \begin{bmatrix}
        R_1 \phi_1+(R_1 + R_2) \sin{(\phi_1 + \phi_2)}\\
        - (R_1 + R_2) \cos{(\phi_1 + \phi_2)}
    \end{bmatrix}, \nonumber\\
    & \uvec{r}_{p_2} = \uvec{r}_{s_2} + \uvec{r}_{p_2/s_2} \nonumber \\
     &  =\begin{bmatrix}
       R_1 \phi_1+ (R_1 + R_2) \sin{(\phi_1 + \phi_2)} + r_2 \sin{(\phi_2 + \theta_2)}\\
        - (R_1 + R_2) \sin{(\phi_1 + \phi_2)} - r_2 \cos{(\phi_2 + \theta_2)}
    \end{bmatrix} 
    \label{EQ:positionvector}
\end{align} 
The derivative of positions vectors (\ref{EQ:positionvector}) is obtained by considering the rotation of the disk on the plane along $x$ axes as
\begin{align}
&\uvec{V}_{s_1}  = \omega_{s_1} R_1 \uvec{i},\nonumber\\
& \uvec{V}_{s_2}  =  \uvec{V}_{s_1} + \uvec{V}_{s_2/s_1}\nonumber\\
 &=  \begin{bmatrix}
        \dot{\phi_1} R_1 + (\dot{\phi_1} + \dot{\phi_2}) (R_1 + R_2) \cos{(\phi_1 + \phi_2)} \\
        (\dot{\phi_1} + \dot{\phi_2}) (R_1 + R_2) \sin{(\phi_1 + \phi_2)}
    \end{bmatrix}\nonumber\\ 
&\uvec{V}_{p_1} = \uvec{V}_{s_1} + \uvec{V}_{p_1/s_1}\nonumber\\
& =  \begin{bmatrix}
        \dot{\phi_1} R_1 + r_1 (\dot{\phi_1} + \dot{\theta_1}) \cos{(\phi_1 + \theta_1)}\\
        r_1 (\dot{\phi_1} + \dot{\theta_1}) \sin{(\phi_1 + \theta_1)}
    \end{bmatrix},\nonumber\\
      &     \uvec{V}_{p_2} = \uvec{V}_{s_2} + \uvec{V}_{p_2/s_2}= \nonumber\\ 
   & \Big[
        \dot{\phi_1} R_1 + (\dot{\phi_1} + \dot{\phi_2}) (R_1 + R_2) \cos{(\phi_1 + \phi_2)} + r_2 (\dot{\phi_2} + \dot{\theta_2}) \nonumber \\ 
        &\cos{(\phi_2+ \theta_2)},\;
         (\dot{\phi_1} + \dot{\phi_2}) (R_1 + R_2) \sin{(\phi_1 + \phi_2)} \nonumber\\&+ r_2 (\dot {\phi_2} + \dot{\theta_2}) \sin{(\phi_2 + \theta_2)}
    \Big]^T,
\end{align}
where $\omega_{s_1}$ is equal to the rotation of the outer body $\dot{\phi_1}$. 

Now, the energy equations define the total kinetic and potential energy of the system which has been formulated for each body and its pendulum from the robot dynamics. Each body's energy equation is presented below. The total energy is considered with inertia and body linear momentum for the first disk and pendulum are as follows
\begin{align}
    E_1 &= \frac{1}{2} m_p ||\uvec{V}_{p_1}||^2 + \frac{1}{2} I_p ||\bm{\omega}_{p_1}||^2  + \frac{1}{2} m_s ||V_{s_1}||^2 \notag \\
    &  + \frac{1}{2} I_s ||\bm{\omega}_{s_1}||^2 + U_1
\end{align}
where $\parallel - \parallel$ is the norm operation and 
\begin{align*}
   & ||\uvec{V}_{p_1}||^2 = 
        \left(R_1^2 + 2R_1 r_1 \cos{(\phi_1 + \theta_1)} + r_1^2\right)  \dot{\phi_1}^2 +
        r_1^2 \dot{\theta_1}^2 \\ &+
        \left(2 R_1 r_1 \cos{(\phi_1 + \theta_1)} + 2r_1^2\right)\dot{\phi_1}\dot{\theta_1},\;\;\\
   & ||\bm{\omega}_{p_1}||^2 = \dot{\theta_1}^2, \;\;||\uvec{V}_{s_1}||^2 = \dot{\phi_1}^2 R_1^2,\;\;
    ||\bm{\omega}_{s_1}||^2 = \dot{\phi_1}^2, \nonumber\\
    &U_{1} = - m_p r_1 g \cos{(\phi_1 + \theta_1)}
,\end{align*}
The second body is similar to the first disk and is computed with the following energy equations of
\begin{align}
    E_2 &= \frac{1}{2} m_p ||\uvec{V}_{p_2}||^2 + \frac{1}{2} I_p ||\bm{\omega}_{p_2}||^2 + \frac{1}{2} m_s ||\uvec{V}_{s_2}||^2 \notag \\
    &\quad  + \frac{1}{2} I_s ||\bm{\omega}_{s_2}||^2 + U_2,
\end{align}
where
\begin{align*}
    &\| \uvec{V}_{p_2}\|^2 = \dot{\phi_1}^2 R_1^2 + (\dot{\phi_1} + \dot{\phi_2})^2 (R_1 + R_2)^2  + r_2^2 (\dot{\phi_2} + \dot{\theta_2})^2 \\
    &\quad + 2 \dot{\phi_1} R_1 (\dot{\phi_1} + \dot{\phi_2}) (R_1 + R_2) \cos{(\phi_1 + \phi_2)} \\
    &\quad + 2 \dot{\phi_1} R_1 r_2 (\dot{\phi_2} + \dot{\theta_2}) \cos{(\phi_2 + \theta_2)} \\
    &\quad + 2 (\dot{\phi_1} + \dot{\phi_2}) (R_1 + R_2) r_2 (\dot{\phi_2} + \dot{\theta_2}) \\
    &\quad \times \cos{(\phi_1 + \phi_2)} \cos{(\phi_1 - \theta_2)}, \\
   & \| \uvec{V}_{s_2} \| ^2 =  R_1^2 + (R_1 + R_2)^2 + 2R_1 (R_1 + R_2) \\
   &\cdot \cos{(\phi_1 + \phi_2)} \dot{\phi_1}^2+
        (R_1 + R_2)^2 + r_2^2\dot{\phi_2}^2  \\
        &+2(R_1 + R_2) r_2 \cos{(\phi_2 + \theta_2)} \dot{\phi_1}\dot{\phi_2}, \\
   &\| \bm{\omega}_{p_2} \|^2 = \dot{\theta_2}^2, \; \| \bm{\omega}_{s_2} \|^2 = \dot{\phi_2}^2, \nonumber\\
   &U_{2} = -m_p g \Big[ (R_1 + R_2) \sin{(\phi_1 + \phi_2)} + r_2 \cos{(\phi_2 + \theta_2)}\Big]. \\
&- m_s g \Big[ (R_1 + R_2) \cos{(\phi_1 + \phi_2)} \Big].
\end{align*}
Finally, we can find the total energy $E$ of the system as:
\begin{align}
   E = E_1 + E_2,
\end{align}
We also incorporate dissipation functions $P$ for both the pendulums and rolling disks to model surface friction and prevent unbounded energy conservation within the system as 
\begin{equation}
    P= \frac{1}{2} \delta_{\theta_1} \dot{\theta}^2_1 + \frac{1}{2} \delta_{\theta_2} \dot{\theta}^2_2 + \frac{1}{2} \delta_{\phi_1} \dot{\phi}^2_1 + \frac{1}{2} \delta_{\phi_2} \dot{\phi}^2_2,
\end{equation}
where are $\bm\delta=\left[\delta_{\theta_1},\delta_{\theta_2},\delta_{\phi_1},\delta_{\phi_2}\right]$ the constants representing dynamic friction. 

The equations represent the torque experienced upon the pendulum and the outer bodies of the disk modules. Therefore using the energy equation to formulate the Euler-Lagrange Equations where $q_i$ represents the variable in question.
\begin{equation}
    \frac{\partial}{\partial t} \Big( \frac{\partial E}{\partial \dot{q_i}} \Big) - \frac{\partial E}{\partial q_i} + \frac{\partial P}{\partial \dot{q}_i}= \tau_{q_i},
    \label{eq:lagranguationEuler}
\end{equation}
where $\tau_{q_i}=\tau_{mag,i}+\tau_{m,i}$ which $\tau_{mag,i}$ and $\tau_m,i$ are the magnetic torque created between permanent magnets located on end-effector of the pendulum and motor torque, respectively. Therefore the equation for the corresponding torques and accelerations can be represented in the form:
\begin{align}
    \uvec{M}(\uvec{q})\ddot{\uvec{q}} + \uvec{C}(\uvec{q},\dot{\uvec{q}})\dot{\uvec{q}} + \uvec{G}(\uvec{q}) = \bm{\tau},
    \label{Eq:Finalmotioneq}
\end{align}
where \( \mathbf{q} \) represents the state vector. The matrix of coefficients can presented as:
\begin{equation}
    \begin{bmatrix}
        a_{11} & 0 & a_{13} & 0 \\
        0 & a_{22} & a_{23} & a_{24} \\
        a_{31} & a_{32} & a_{33} & a_{34} \\
        0 & a_{42} & a_{43} & a_{44}
    \end{bmatrix}
    \begin{bmatrix}
        \ddot{\theta_1} \\ \ddot{\theta_2} \\ \ddot{\phi_1} \\ \ddot{\phi_2}
    \end{bmatrix} +
    \begin{bmatrix}
        x_1 \\ x_2 \\ x_3 \\ x_4
    \end{bmatrix}+
    \begin{bmatrix}
        y_1 \\ y_2 \\ y_3 \\ y_4
    \end{bmatrix} = 
    \begin{bmatrix}
        \tau_{\theta_1} \\ \tau_{\theta_2} \\ \tau_{\phi_1} \\ \tau_{\phi_2}
    \end{bmatrix}
    \label{Eq:Matrixdetailstorquemodel}
\end{equation}
where
{\footnotesize
\begin{align*}
    &a_{11} = m_p r_1^2 + I_p, \quad a_{13} = r_1^2 + R_1 r_1 \cos{(\phi_1 + \theta_1)}, a_{22} = m_p r_2^2 + I_p, \\
    &a_{23} = m_p \Big( r_2 + \cos{(\phi_1 + \theta_2)} + (R_1 + R_2) r_2 \cos{(\phi_1 + \phi_2)}\cos{(\phi_1 + \theta_2)} \Big) \\
    &a_{24} = m_p (R_1 + R_2) r_2 \cos{(\phi_1 + \phi_2)} \cos{(\phi_1 + \theta_2)} \\
    &a_{31} = m_p \Big(R_1 r_1 \cos{(\phi_1 + \theta_1)}  - R_1(R_1 + R_2)\cos{(\phi_1 + \phi_2)} \Big) \\
    & + m_s \Big(R_1(R_1 + R_2) \cos{(\phi_1 + \phi_2)} \Big) \\
    &a_{32} = m_p (R_1 + R_2)r_2 \Big( \cos{(\phi_2 + \theta_2)} - \sin{(\phi_1 + \phi_2)} \Big) \\
    &- m_s R_1 (R_1 + R_2) \sin{(\phi_1 + \phi_2)} \\
    &a_{33} = m_s R_1^2 + I_s + m_p \Big[2R_1^2 + (R_1 + R_2)^2 \\
    &+ R_1(R_1 + R_2)\cos{(\phi_1 + \phi_2)} \Big] \\
    &a_{34} = m_p \Big((R_1 + R_2)r_2 \cos{(\phi_1 + \phi_2)}- (R_1 + R_2)\cos{(\phi_1 - \theta_2)} \Big) \\
    &+ m_s \Big((R_1 + R_2)^2 + R_1(R_1 + R_2) \Big)  \cos{(\phi_1 + \phi_2)} \\
    &a_{42} = m_p r_2^2 + m_p (R_1 + R_2) r_2  \cos{(\phi_1 + \phi_2)} \cos{(\phi_1 - \theta_2)} \\
    &a_{43} = (m_p + m_s)(R_1 + R_2)^2 + m_p R_1 (R_1 + R_2) \\
    & \times \cos{(\phi_1 + \phi_2)} + m_p (R_1 + R_2) r_2  \cos{(\phi_1 + \phi_2)} \cos{(\phi_1 - \theta_2)} \\
    &a_{44} = (m_p + m_s)(R_1 + R_2)^2 + m_p r_2^2 + I_s \\
    & + m_p (R_1 + R_2) r_2 \cos{(\phi_1 + \phi_2)} \cos{(\phi_1 - \theta_2)}
\end{align*}}
which, also have
{\footnotesize
\begin{align*}
    &x_1 =  R_1 r_1 \sin{(\phi_1 + \theta_1)} \left[ (m_p - 1) \dot{\phi_1} (\dot{\phi_1} + \dot{\theta_1}) \right] +\delta_{\theta_1} \dot{\theta}_1  \\
    &x_2 =  - m_p \left[\dot{\phi_1} \sin{(\phi_1 + \theta_2)} (\dot{\phi_1} + \dot{\theta_2}) \right] +\delta_{\theta_2} \dot{\theta}_2\\
    & \left. + (R_1 + R_2) r_2 (\dot{\phi_1} + \dot{\phi_2}) \sin{(2\phi_1 + \phi_2 + \theta_2)} (\dot{\phi_1} + \dot{\phi_2}) \right] \\
    & - m_p \left[\dot{\phi_1} R_1 r_1 (\dot{\phi_2} + \dot{\theta_2}) \sin{(\phi_2 + \theta_2)} \right. \\
    & \left. + (\dot{\phi_1} + \dot{\phi_2}) (\dot{\phi_2} + \dot{\theta_2}) (R_1 + R_2) \sin{(\phi_1 - \theta_2)} \right] \\
    &x_3 =  -R_1 r_1 \sin{(\phi_1 + \theta_1)} (\dot{\phi_1} + \dot{\theta_1})^2 \\
    & + m_p R_1 r_1 \sin{(\phi_1 + \theta_1)} \left(\dot{\phi_1}^2 + \dot{\phi_1} \dot{\theta_1}\right) \\
    & - m_p (\dot{\phi_1} + \dot{\phi_2}) (R_1 + R_2) r_2 (\dot{\phi_2} + \dot{\theta_2})  \sin{(2\phi_1 + \phi_2 - \theta_2)} \\
    & - (R_1 + R_2) r_2 \Big[ (\dot{\phi_1} + \dot{\phi_2})^2 (\dot{\phi_2} + \dot{\theta_2}) \sin{(\phi_1 - \theta_2)} \cos{(\phi_1 + \phi_2)} \Big] \\
    & + m_s R_1 (R_1 + R_2) \sin(\phi_1 + \phi_2) \left( \dot{\phi_1}^2 + \dot{\phi_1} \dot{\phi_2} \right)+ \delta_{\phi_1} \dot{\phi}_1 \\
    &x_4 =  - (m_p + m_s) R_1 (R_1 + R_2) \sin{(\phi_1 + \phi_2)} (\dot{\phi_1}^2 + \dot{\phi_1} \dot{\phi_2}) \\
    & - m_p (R_1 + R_2) r_2 \sin{(\phi_1 + \phi_2)}  (\dot{\phi_1} + \dot{\phi_2}) (\dot{\phi_1} + 2 \dot{\phi_2} + \dot{\theta_2}) \\
    & - m_p R_1 r_2 \sin{(\phi_2 + \theta_2)} (\dot{\phi_1} + \dot{\phi_2}) \dot{\phi_1} + m_p R_1 (R_1 + R_2) \dot{\phi_1} \\
    & \times (\dot{\phi_1} + \dot{\phi_2}) \sin{(\phi_1 + \phi_2)} + m_p R_1 r_2 \dot{\phi_1} (\dot{\phi_2} + \dot{\theta_2}) \sin{(\phi_2 + \theta_2)} \\
    & - m_p (R_1 + R_2) r_2 (\dot{\phi_2} + \dot{\theta_2}) (\dot{\phi_1} + \dot{\phi_2}) \cos{(\phi_1 - \theta_2)} \sin{(\phi_1 + \phi_2)} \nonumber\\
    & + \delta_{\phi_1} \dot{\phi}_1
\end{align*}}
{\footnotesize
\begin{align*}
    y_1 = & - m_p g \Big( r_1  \sin{(\phi_1 + \theta_1)} + (R_1 + R_2) \cos{(\theta_1)} \Big) \\
    y_2 = & - m_p g r_2 \cos{(\phi_2 + \theta_2)}, y_3 =   m_p r_1 g \sin{(\phi_1 + \theta_1)} \\
    & - (R_1 + R_2) g \Big[ -m_p \cos{(\phi_1 + \phi_2)} + m_s \sin{(\phi_1 + \phi_2)} \Big] \\
    y_4 = & - m_p g r_2 \sin{(\phi_2 + \theta_2)}  + \Big[ (m_s - m_p) g (R_1 + R_2) \sin{(\phi_1 + \phi_2)} \Big] \\
    & - m_p g (R_1 + R_2) \cos{(\phi_1 + \phi_2)}. \\
\end{align*}}
It is important to note that, based on insights from our experimental studies, we observed that the motor exhibits a two-sided torque reaction affecting both the disk and pendulum; hence, we define the motor torques in (\ref{Eq:Matrixdetailstorquemodel}) as $\tau_1 = \tau_{\theta_1} = -\tau_{\phi_1}$ and $\tau_2 = \tau_{\theta_2} = -\tau_{\phi_2}$.

\subsection{Magnetic Coupling Model}
\begin{figure}[t!]
    \centering
    \includegraphics[width=0.38\linewidth]{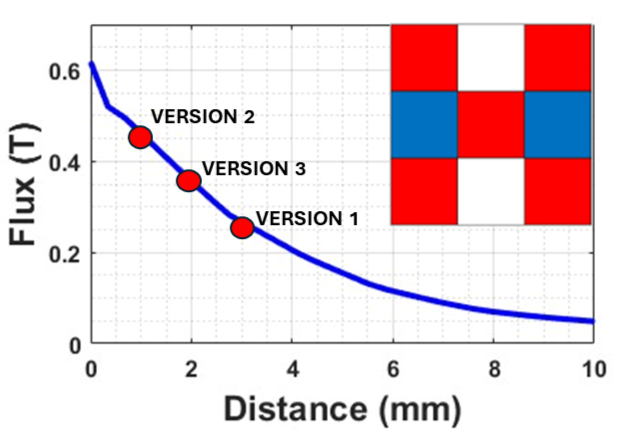}
    \includegraphics[width=0.38\linewidth]{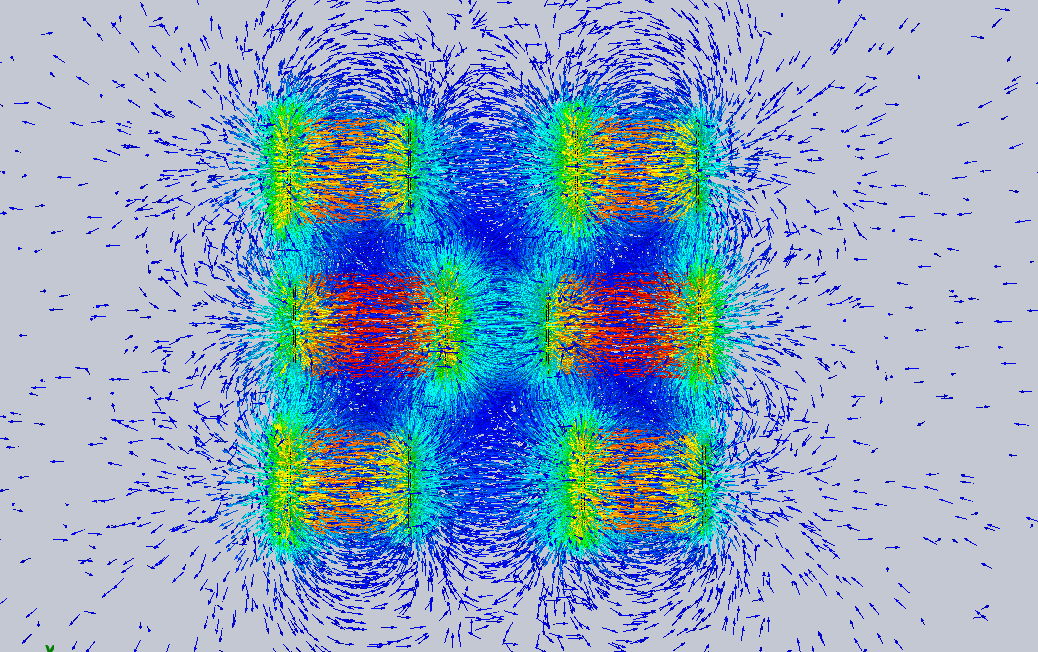}\\
    \hspace{.3cm}(a) \hspace{3cm} (b)
    \caption{a) Simulation results for magnetic flux versed different distances Blue = Positive / Red = Negative Pole) b) Magnetic flux formation}\label{fig:final arrangment}
\end{figure}
To calculate the magnetic force between the pendulums' coupling permanent magnets (shown in Fig. \ref{fig:robot_design} and Fig. \ref{fig:final arrangment}), we model the magnetic flux density \( B_{mag} \) as a function of the distance \( p_m \) between them based on our experiments. We assume a linear approximation for the magnetic flux density, which decreases as the distance between the pendulums increases, up to a maximum effective distance \( P_{max} \); hence, The magnetic flux density, \( B_{mag} \), is expressed by the following equation:
\begin{equation}
B_{mag} = 
\begin{cases} 
B_{max} \cdot \left(1 - \frac{p_m}{P_{max}}\right), & \text{if } p_m \leq P_{max} \\
0, & \text{otherwise}
\end{cases}
\label{eq:magneticflux}
\end{equation}
where \( B_{max} \) is the maximum magnetic flux density which is calculated with Solidworks EMWorks accurate modelled measurements \cite{Ollie2024towards} (example shown in Fig. \ref{fig:final arrangment}-a), and \( p_m \) is the separation distance between the magnets. Please note that $
p_m = \sqrt{(r_{p1_i} - r_{p2_i})^2 + (r_{p1_j} - r_{p2_j})^2}$ where \( r_{p1_i} \) and \( r_{p1_j} \) are the \( i \)- and \( j \)-components of the position of the first pendulum, and \( r_{p2_i} \) and \( r_{p2_j} \) are the corresponding components of the second pendulum's position.

We then compute the magnetic force \( F_{mag} \) using (\ref{eq:magneticflux}) between the magnets as follows 
\begin{equation}
F_{mag} = \frac{B_{mag}^2 \cdot A}{2  \mu_0},
\end{equation}
where \( A \) is the effective area of magnetic interaction in square meters where a configuration of seven magnets is the cap area as shown in Fig. \ref{fig:final arrangment} ($A=$ 0.0025 m$^2$), and \( \mu_0 = 1.257 \times 10^{-6} \, \text{T} \text{m/A} \) is the magnetic permeability of free space. Finally, the created torque between the two pendulum bodies is formulated as $\tau_{mag,i}$ for  $\tau_{q,i}$ in (\ref{eq:lagranguationEuler}).

\subsection{Designed PD Controller}\label{sec:PD}
To achieve stable coupling of the magnetic interaction between the pendulums, we implement a Proportional-Derivative (PD) controller with multi-input capabilities. This controller is designed to address two primary objectives: (i) control the true angular location of the pendulums, denoted by \( \psi \), to ensure balanced oscillations and stable magnetic coupling; and (ii) maintain the desired angular position of the disks, represented by \( \varphi \), to support the pendulums’ stabilization. In robotic manipulation, non-prehensile manipulation refers to the ability to move objects without the use of gripping or holding, which is particularly relevant in systems involving rolling bodies, such as two disks that are in contact with a moving pendulum \cite{ruggiero2018nonprehensile}. These types of systems are characterized as under-actuated, meaning they have fewer actuators than degrees of freedom. As such, the control and stability of these systems present unique challenges that distinguish them from classical under-actuated systems, such as two-link robotic arms or two disks rotating under external controls. The friction between the disks and the ground plays a critical role, as it can both assist and hinder motion. These interactions necessitate innovative control strategies to stabilize the system effectively.

For both pendulum-disk pairs, the control input vector \( \mathbf{u} = [u_1, u_2]^T \) can be expressed in matrix-vector form as 
\begin{equation}
\mathbf{u} = \mathbf{K}_p 
\begin{bmatrix} 
\psi_1 - \theta_{1d} \\ 
 \psi_2 - \theta_{2d} \\ 
\varphi_1 - \varphi_{1d} \\ 
\varphi_2 - \varphi_{2d} 
\end{bmatrix} 
+ \mathbf{K}_d 
\begin{bmatrix} 
\dot{\theta}_1 - \dot{\theta}_{1d} \\ 
\dot{\theta}_2 - \dot{\theta}_{2d} \\ 
\dot{\varphi}_1 - \dot{\varphi}_{1d} \\ 
\dot{\varphi}_2 - \dot{\varphi}_{2d} 
\end{bmatrix} = \mathbf{K}_p \uvec{q}_c + \mathbf{K}_d \dot{\uvec{q}}_c
\end{equation}
where $\psi_1=\theta_1-\phi_1$, $\psi_2=\theta_2-\phi_2$, \( \mathbf{K}_p \) and \( \mathbf{K}_d \) are the proportional and derivative gain matrices, defined as 
\begin{align}
\mathbf{K}_p &= 
\begin{bmatrix}
k_{p\theta_1} & 0 & k_{p\varphi_1} & 0 \\ 
0 & k_{p\theta_2} & 0 & k_{p\varphi_2} 
\end{bmatrix}, \nonumber\\
\mathbf{K}_d &= 
\begin{bmatrix}
k_{d\theta_1} & 0 & k_{d\varphi_1} & 0 \\ 
0 & k_{d\theta_2} & 0 & k_{d\varphi_2} 
\end{bmatrix}
\end{align}
where \( k_{p\theta} \) and \( k_{d\theta} \) terms in each row of \( \mathbf{K}_p \) and \( \mathbf{K}_d \) primarily influence the pendulum's true angular location, ensuring that the magnetic interaction remains stable and the pendulums are balanced at their desired angles \( \theta_{1d} \) and \( \theta_{2d} \).

In analysing the stability of our PD controller, we propose a generic stability criterion based on asymptotic stability. We can re-present the closed-loop dynamics from (\ref{Eq:Finalmotioneq}) governed by the PD controller can be expressed as:
\begin{equation}
\dot{\mathbf{q}} = \mathbf{f}(\mathbf{q}) + \mathbf{g}(\mathbf{q})\mathbf{u},
\end{equation}
where \( \mathbf{f}(\mathbf{q}) \) encapsulates the system's inherent non-linear dynamics, and \( \mathbf{g}(\mathbf{q})\mathbf{u} \) denotes the influence of the control input on the state.

To ensure local boundedness for stability, we require that the control input \( \mathbf{u} \) remains within specific limits, typically defined by the maximum actuator capabilities. If we denote the control input constraints as \( \| \mathbf{u} \| \leq M \), where \( M \) is a positive constant representing the maximum allowable input, we can establish that the state trajectories are locally bounded as
\begin{equation}
\| \dot{\mathbf{q}} \| \leq \| \mathbf{f}(\mathbf{q}) \| + \| \mathbf{g}(\mathbf{q}) \| \cdot M,
\end{equation}
\begin{figure}[t!]
    \centering
    \includegraphics[width=1\linewidth]{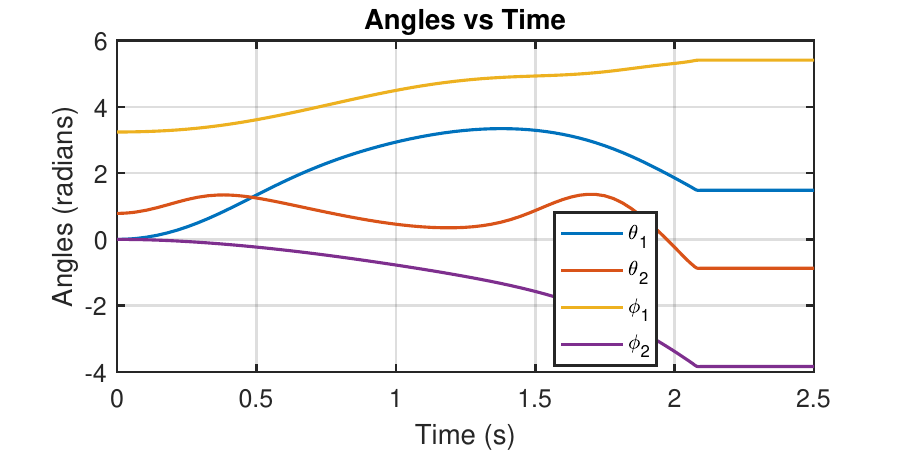}
    \includegraphics[width=3.5in]{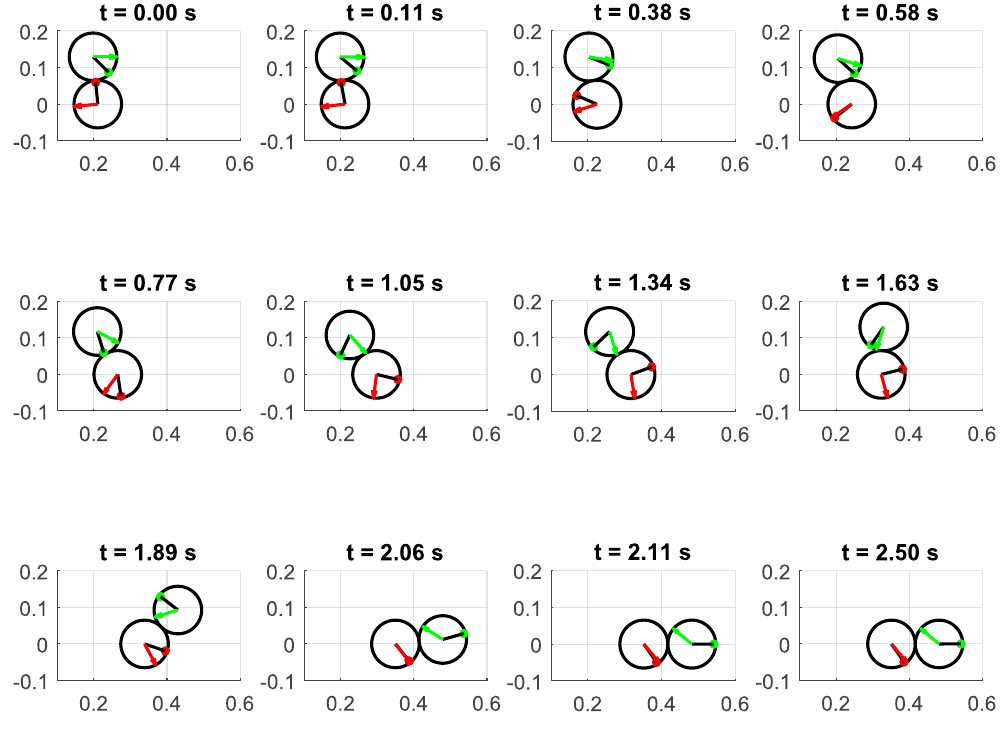}
    \caption{Simulation results demonstrating the system left to free fall}
    \label{fig:freefall}
\end{figure}

To analyse and keep the system in a stable region after confirming the boundedness, we can define the Lyapunov function of the whole system including the controller as:
\begin{equation}
\uvec{V}(\mathbf{q}) = \frac{1}{2} \uvec{I}_{2\times2} \uvec{E}^2 + \frac{1}{2} \mathbf{K}_p \uvec{q}^2_c + \mathbf{K}_d \dot{\uvec{q}}^2_c
\label{Eq:Lypanuovtotal}
\end{equation} 
By taking the derivative along the trajectories of the system (\ref{Eq:Lypanuovtotal}) can be computed as:
\begin{equation}
\dot{\uvec{V}}=  \uvec{I}_{2\times2} \uvec{E}\dot{\uvec{E}}+ \mathbf{K}_p \uvec{q}_c \dot{\uvec{q}}_c+ \mathbf{K}_d \dot{\uvec{q}}_c \ddot{\uvec{q}}_c
\end{equation}
If \( \dot{\uvec{V}}(\mathbf{q}) \)$\ll 0$ exhibits asymptotic stability as $t\rightarrow \infty$ indicating that the system will return to equilibrium, albeit at varying rates depending on the initial conditions and the specific non-linear characteristics of the dynamics involved. If we can show that the derivative \( \dot{\uvec{V}}(\mathbf{q}) \) is negative definite or at least negative semi-definite for the closed-loop dynamics, we can conclude that the equilibrium point is stable. Thus, the gains of the controller are chosen accordingly to be in a stable region.

\begin{figure}[t!]
    \centering
    \includegraphics[width=1\linewidth]{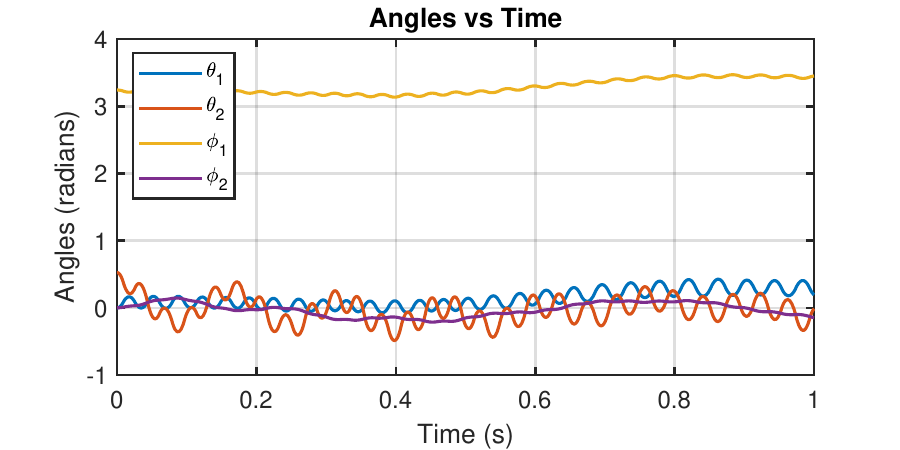}
    \includegraphics[width=3.4in]{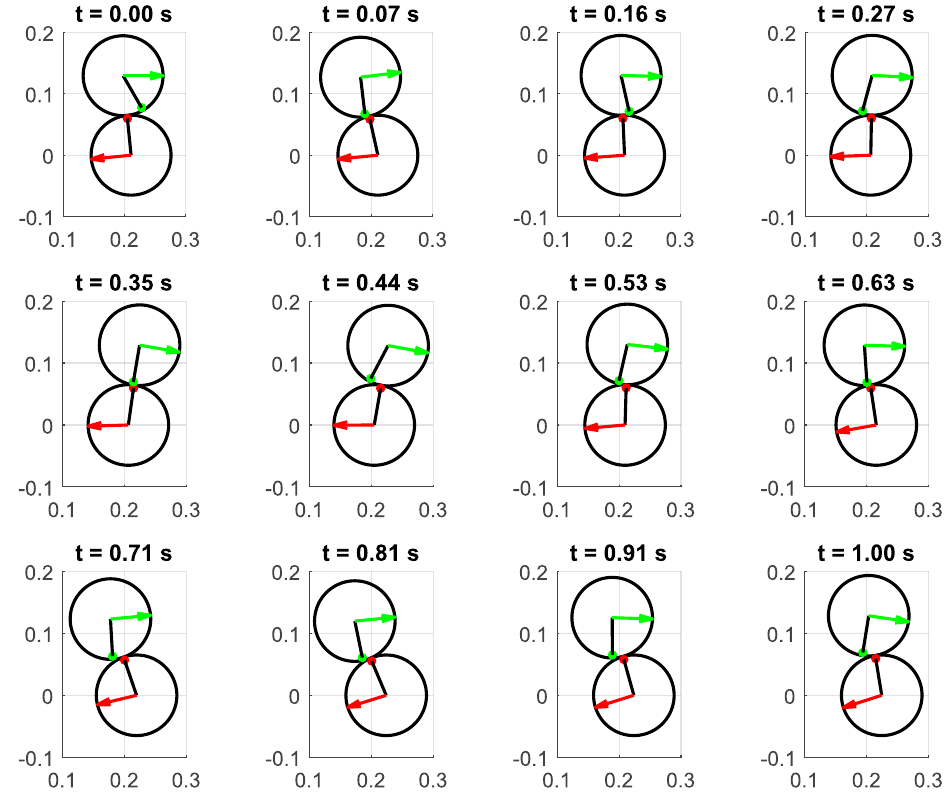}
    \caption{Simulation results for balancing PD controller}
    \label{fig:PD_simulation}
\end{figure}

\section{Simulation Studies}\label{sec:simulation}
\begin{figure}[t!]
    \centering
    \includegraphics[width=1\linewidth]{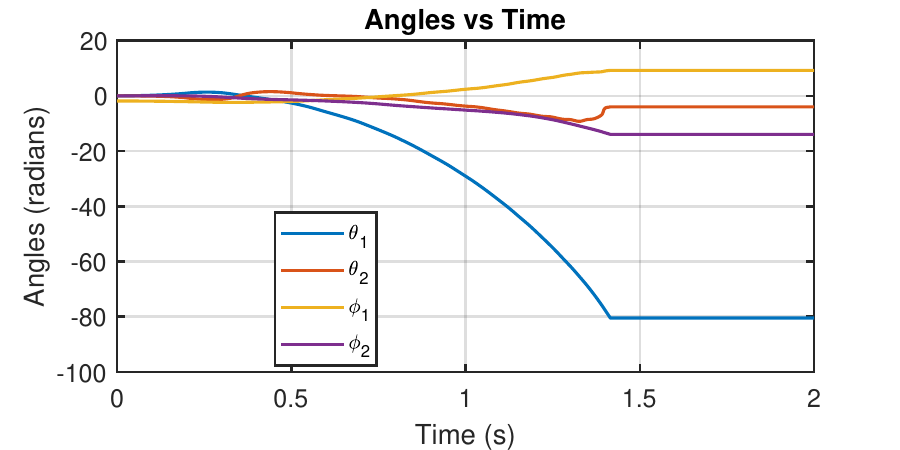}
    \includegraphics[width=3.3in]{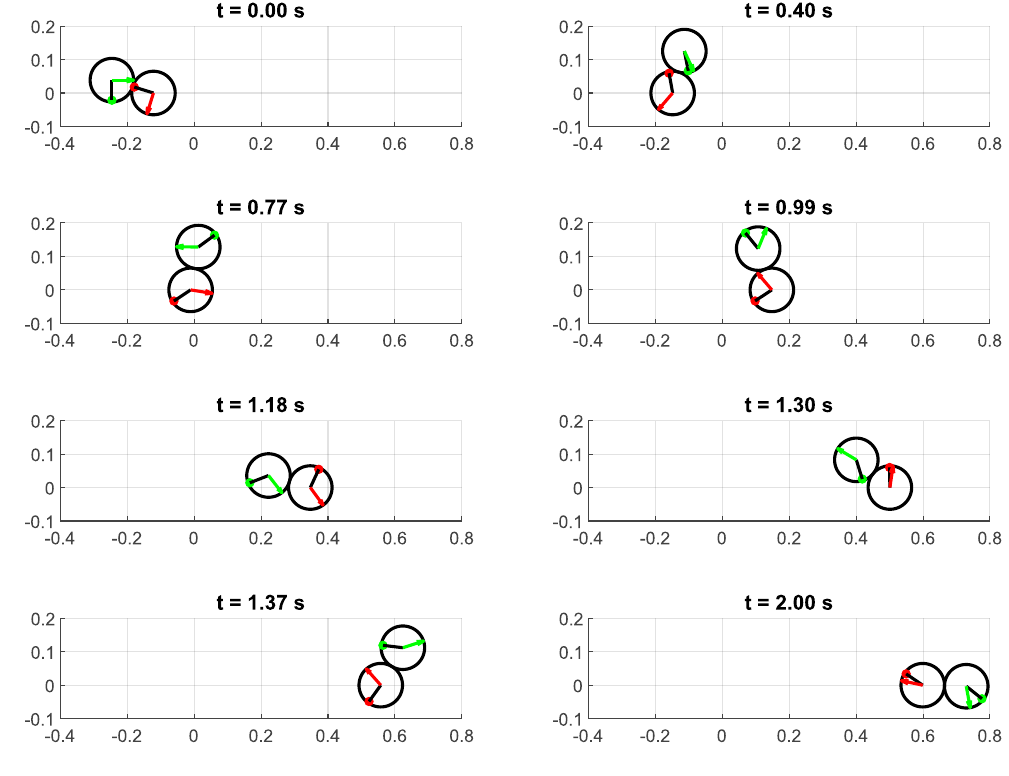}
    \caption{Simulation results for a lifting rolling motion}
    \label{fig:rolling}
\end{figure}
In this section, we analyze three scenarios: first, we examine the free-fall motion of the two disks, then we assess the effect of a PD controller designed to balance the motion of the two disks, and finally, we evaluate another new nonprehensile lifting capability of the system. Please note that for these scenarios based on studies on physical systems and their surface friction studies the damping coefficients are determined as $\bm\delta=[0.02\;\;0.02\;\;0.07\;\;0.06]^T$. The rest of the real physical parameters are taken from the designed real reconfigurable disk robot \cite{Ollie2024towards} as shown in Table \ref{tab:parameters}.

At first, we need to confirm the physics is working properly in the modelled system with the initial condition of \( y_0 = \begin{bmatrix} \theta_1 & \theta_2 & \phi_1 & \phi_2 & \dot{\theta}_1 & \dot{\theta}_2 & \dot{\phi}_1 & \dot{\phi}_2 \\ \end{bmatrix} \) = $ \begin{bmatrix}
0^\circ & 30^\circ & 185^\circ & 0^\circ & 0 & 0 & 0 & 0 \\ \end{bmatrix}$. The disks rest on each other, with the upper pendulum offsetting the balance point. Without control inputs $\uvec{u}$, the system quickly topples and falls due to unopposed gravitational forces, as shown in Fig. \ref{fig:freefall}, which confirms the system works naturally.

For a PD-controlled case with challenging initial conditions similar to the free-fall scenario, \( y_0 = \begin{bmatrix} 0^\circ & 30^\circ & 180^\circ & 185^\circ & 0 & 0 & 0 & 0 \end{bmatrix} \), we select the controller gains as 
\begin{equation*}
    \mathbf{K}_p = 
    \begin{bmatrix}
        20 & 0 & 0.3 & 0 \\ 
        0 & 30 & 0 & 0.7 
    \end{bmatrix}, \mathbf{K}_d = 
    \begin{bmatrix}
        0.1 & 0 & 0.5 & 0 \\ 
        0 & 0.2 & 0 & 0.75 
    \end{bmatrix}
\end{equation*}
While the PD controller, as described in Section \ref{sec:PD}, generally balances the system, some minor instability due to control-input oscillations is observed. In future work, a more advanced strategy, such as sliding mode control, could be explored to better handle balancing dynamics and potential external disturbances.

Next, as reconfigurable rolling robots are intended for configurations with multiple rolling bodies, direct stacking is impractical, necessitating a \textit{lifting motion} to place one module atop another. Through extensive testing, we observed unique motion behaviors (Fig. \ref{fig:rolling}) during simulation. Specifically, by tuning the PD gains, the bottom disk successfully lifted and briefly balanced the second disk before toppling—a potentially valuable movement pattern for reconfiguration. The specific gains and initial conditions for these tests are as $
    y_0 = 
    \begin{bmatrix}
        0^\circ & 0^\circ & 0^\circ & -107.19^\circ & 0 & 0 & 0 & 0 \\ 
    \end{bmatrix}$
    and 
\begin{equation*}
    \mathbf{K}_p = 
        \begin{bmatrix}
        1.3 & 0 & 0.3 & 0 \\ 
        0 & 2.5 & 0 & 1.8 
        \end{bmatrix},
        \mathbf{K}_d = 
        \begin{bmatrix}
        0.1 & 0 & 0.5 & 0 \\ 
        0 & 0.2 & 0 & 0.35 
    \end{bmatrix}
\end{equation*}
This motion pattern is intriguing, as the bottom disk effectively acts as an actuator through frictional interactions between modules. Mastering this motion could enable a novel form of locomotion for reconfigurable systems, presenting a new nonprehensile control challenge that involves mass balancing and slip management without direct physical manipulation.

\section{Conclusion}
This paper presented a simulation study of a novel mobile rolling disk robot designed for non-prehensile manipulation, focusing on the unique control challenges of its internally actuated magnetic-pendulum coupling mechanism. The system's dynamics are driven through Lagrangue-Euler formulation by the interplay of frictional and magnetic effects between modules. Through simulations, we analyzed three scenarios: PD-controlled balancing, free-fall behavior, and a novel lifting motion. Results show that, without control, the system quickly topples under gravity, while the PD controller effectively maintains stability by continuously adjusting the top disk's position to counteract imbalances, though some oscillatory behavior was noted.

The simulations validated the system’s dynamics under gravitational forces, showing that certain initial conditions led to the toppling of the upper body, while others enabled the bottom disk to lift and momentarily balance the second disk. This behavior indicates a potential reconfiguration motion pattern, presenting a new nonprehensile manipulation problem as lifiting. These findings underscore the need for advanced control strategies to enhance stability and manage complex behaviors in modular robotics.

In future work, we aim to implement a sliding-mode controller to address challenging scenarios like lifting with slippage and experiment with physical robots. Additionally, we plan to develop a spherical, reconfigurable form of the robot to tackle practical applications.






\section{Acknowledgment}
This work was supported by the Royal Society research grant under Grant \text{RGS\textbackslash R2\textbackslash 242234}.


 \bibliographystyle{ieeetr}
 \bibliography{references}

\end{document}